\documentclass[11pt,a4paper]{article}
\usepackage[hyperref]{emnlp2018}
\usepackage{times}
\usepackage{latexsym}

\usepackage{url}

\usepackage{graphicx}
\usepackage{amsmath}
\usepackage{amsfonts}
\usepackage{algorithm}
\usepackage{algorithmic}

\usepackage{float}
\usepackage{multirow}
\usepackage{verbatim}
\usepackage{caption}
\usepackage{subcaption}

\usepackage{xcolor}

\newcommand{\RN}[1]{%
	\textup{\uppercase\expandafter{\romannumeral#1}}%
}

\aclfinalcopy 

\title{Actor-Critic based Training Framework for Abstractive Summarization}

\author{ Piji Li$^{\dag}$  \ \ Lidong Bing$^{\ddag}$  \ \ Wai Lam$^{\dag}$\\
	$^{\dag}$Department of Systems Engineering and Engineering Management,\\
	The Chinese University of Hong Kong\\
	$^{\ddag}$AI Lab, Tencent Inc., Shenzhen, China\\
	{\tt  $^{\dag}$\{pjli, wlam\}@se.cuhk.edu.hk, $^{\ddag}$lyndonbing@tencent.com}}

\date{}

\begin{document}
	\maketitle
	\begin{abstract}
		We present a training framework for neural abstractive summarization based on actor-critic approaches from reinforcement learning. In the traditional neural network based methods, the objective is only to maximize the likelihood of the predicted summaries, no other assessment constraints are considered, which may generate low-quality summaries or even incorrect sentences. To alleviate this problem, we employ an actor-critic framework to enhance the training procedure. For the actor, we employ the typical attention based sequence-to-sequence (seq2seq) framework as the policy network for summary generation. For the critic, we combine the maximum likelihood estimator with a well designed global summary quality estimator which is a neural network based binary classifier aiming to make the generated summaries indistinguishable from the human-written ones. Policy gradient method is used to conduct the parameter learning. An alternating training strategy is proposed to conduct the joint training of the actor and critic models. Extensive experiments on some benchmark datasets in different languages show that our framework achieves improvements over the state-of-the-art methods. 
	\end{abstract}
	
	\section{Introduction}
	Text summarization, aiming at automatically generating a brief, well-organized summary for an input document, has been studied extensively \cite{nenkova2012survey,yao2017recent}.
	Summarization approaches can be grouped into two classes: extraction-based methods and abstraction-based methods.
	Extractive summarization is to extract the original sentences from the source documents to create a short summary \cite{erkan2004lexrank,min2012exploiting}.
	Since the sentences are extracted from the original documents directly, the summary can guarantee good linguistic quality.
	However, the extracted sentences probably contain some unimportant phrases or clauses. Considering the length constraint of the summary, text space is wasted due to the redundant information. Therefore, some researchers utilize compression strategies to remove the noisy information from the sentences \cite{wang2013sentence}, but the main summarization techniques are still based on extraction.
	
	\begin{figure}[!t]
		\centering
		\small
		\begin{tabular}{l | p{0.75\columnwidth}}
			\hline
			\textbf{Input} & president  's springer spaniel , spot , was old and ailing . \\
			\textbf{Human}& former head usher at white house witnessed first families at their most vulnerable\\
			\textbf{Seq2seq}& \textit{president 's spaniel spaniel spaniel spaniel becomes secretary of s springer spaniel spaniel spaniel spaniel spaniel spaniel}\\
			\hline
			\textbf{Input} & u.s. home resales posted the largest monthly increase in at least \#\# years last month as first-time buyers rushed to take advantage of a tax credit that expires this fall .\\
			\textbf{Human}&july home sales surge more than \# percent	\\
			\textbf{Seq2seq}& \textit{u.s. home resales rose \#.\# \% to \#.\#\#\# bln rate in \#\# years}\\
			\hline
		\end{tabular}
		\caption{\label{fig:front}
			Cases of low-quality summaries generated by the typical attention based seq2seq framework. The first one contains repetitive words. The second one contains lots of noisy symbols (\#). 
		}
		\vspace{-3mm}
	\end{figure}
	
	Recently, with the rapid development of deep learning techniques, especially benefits from the attention based sequence-to-sequence (seq2seq) framework \cite{bahdanau2014neural}, some neural abstractive summarization approaches are proposed and achieve significant improvement \cite{rush2015neural,chopra2016abstractive,tan2017abstractive,zhou2017selective,li2017deep}. \citeauthor{rush2015neural} \shortcite{rush2015neural} propose a neural network based model with local attention modeling for abstractive sentence summarization. \citeauthor{nallapati2016abstractive} \shortcite{nallapati2016abstractive} extend the sentence summarization model by using a hierarchical seq2seq structure to encode the long input content. 
	
	Although the proposed frameworks are able to generate abstractive summaries, after conducting the investigation, we still notice that there are several obvious problems implied in the results. One problem is that typical seq2seq frameworks often generate unnatural summaries consisting of repeated words or phrases. Take the first case in Figure~\ref{fig:front} as an example, the sequence generated by the model contains repeated word ``spaniel'', which is even an incorrect sentence.
	Another problem is that some sentences generated by seq2seq models are not homogeneous with the summaries written by human. For example, some sentences contain lots of unimportant numbers or meaningless symbols, as shown in the second case in Figure~\ref{fig:front}; Some sentences only convey general information, such as ``it's all about you''; Some suffer the problem of Out-Of-Vocabulary (OOV), such as ``$<$unk$>$ the $<$unk$>$ of $<$unk$>$''.
	Intuitively, if we can obtain negative evaluation signals for these low quality summaries and use the signal to guard the training, the summarization performance can be improved. However, maximum likelihood estimation (MLE) cannot handle this issue.
	
	To tackle the above mentioned problems, we propose a training framework based on \textbf{Actor-Critic} (AC) approaches \cite{konda2000actor,bahdanau2017actor} from the area of reinforcement learning \cite{sutton1998reinforcement}.
	We employ the typical attention based seq2seq framework as the actor to conduct the summary generation. For the critic, we combine the maximum likelihood estimator with a well designed global summary quality estimator. The summary quality estimator is a neural network based binary classifier which can be regarded as a discriminator between generated summaries and the ground truth. The aim is to make the generated summaries indistinguishable from the human-written ones. 
	Policy gradient method is used to conduct the parameter learning. An alternating training strategy is designed to jointly train the actor and critic models.
	

	The main contributions of our framework are summarized as follows:
	(1) We propose a training framework for neural abstractive summarization based on actor-critic approaches from the area of reinforcement learning.
	(2) We combine the maximum likelihood estimator with a well designed global summary quality estimator for the critic component. The summary quality estimator is a neural network based binary classifier which aims to make the generated summaries indistinguishable from the human-written ones. 
	(3) We introduce an alternating training method to optimize the actor network and the critic network jointly. Policy gradient method is used to conduct the parameter learning.
	(4) Experimental results on some benchmark datasets in different languages show that our framework achieves better performance than the state-of-the-art models.

	\section{Framework}
	
	\subsection{Overview}
	\begin{figure*}[!t]
		\centering
		\includegraphics[width=1.5\columnwidth]{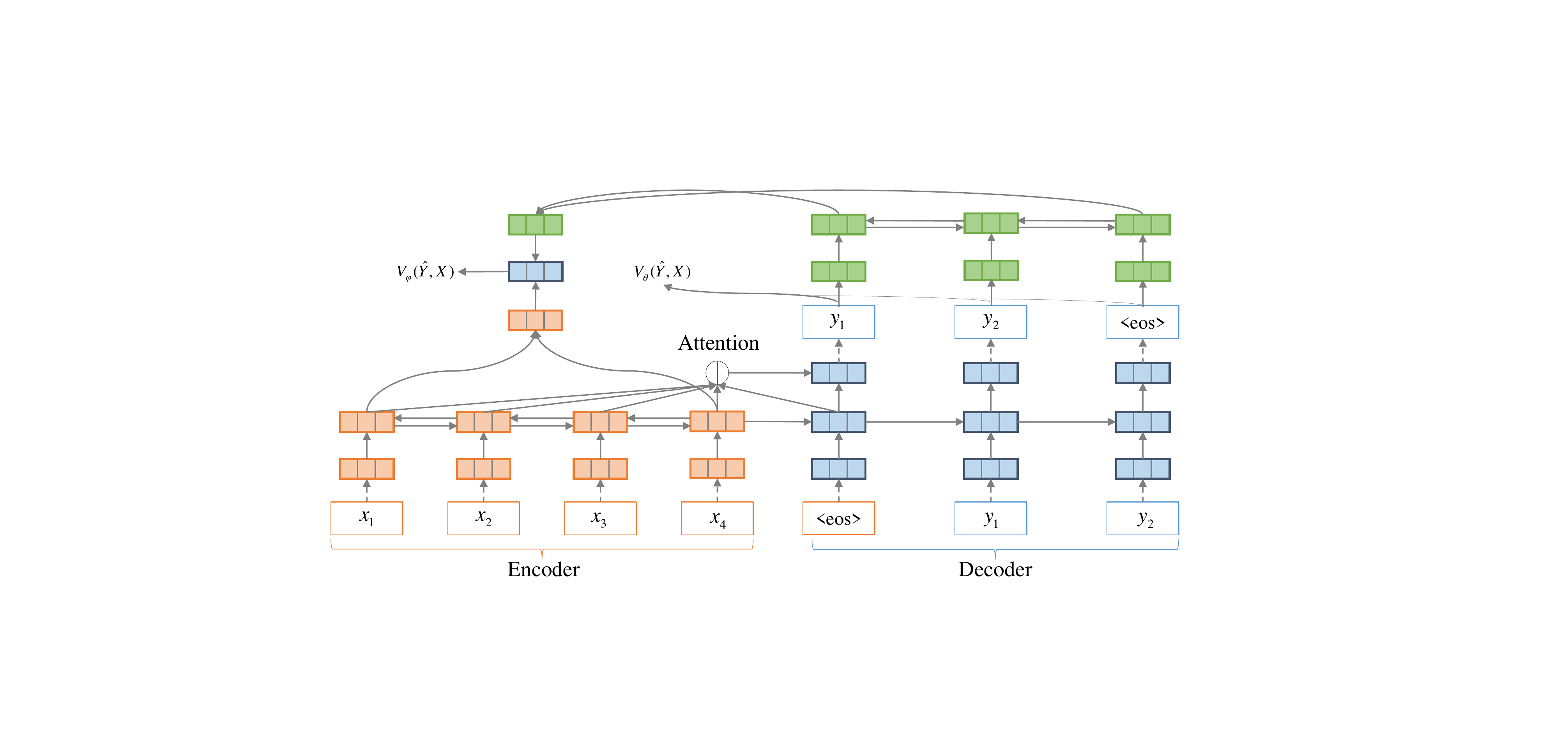}
		\caption{Our actor-critic training framework for abstractive summarization.}
		\label{fig:framework}
		\vspace{-3mm}
	\end{figure*}
	
	Assume that the input is a variable-length sequence ${X} = (\mathbf{x}_1, \mathbf{x}_2, \ldots, \mathbf{x}_m)$ representing the source text.
	The word embedding $\mathbf{x}_t$ is initialized randomly and learned during the optimization process. 
	The output ground truth is also a sequence ${Y} = (\mathbf{y}_1, \mathbf{y}_2, \ldots, \mathbf{y}_n)$.
	We denote the generated summary sequence as $\hat{Y} = (\hat{\mathbf{y}}_1, \hat{\mathbf{y}}_2, \ldots, \hat{\mathbf{y}}_n)$.
	Gated Recurrent Unit (GRU) \cite{cho2014learning} is employed as the basic sequence modeling component for the encoder and the decoder. Global soft attention modeling \cite{bahdanau2014neural} is used to enhance the decoding performance.
	Under our proposed actor-critic framework as shown in Figure~\ref{fig:framework}, we can regard the attention-based seq2seq framework as the policy network $G_\theta$ for the actor, where $\theta$ is the actor parameters. We employ two assessment criteria as the value functions for the critic. Critic \RN{1} ($C_\RN{1}$) is the maximum likelihood estimator. We use the negative log likelihood as the value function $V_{\theta}(\hat{Y}, X)$ for $C_\RN{1}$.
	We also design a global summary quality estimator $V_{\phi}(\hat{Y}, X)$ for Critic II ($C_{\RN{2}}$). $V_{\phi}$ is a binary-class discriminator with parameter $\phi$ that only the ground truth $Y$ is regard as the positive instance, i.e.,  $V_{\phi}(Y, X) \simeq 1$. Policy network for the actor is pre-trained at first, and then we introduce an alternating training method to optimize the actor and critic jointly. Policy gradient method is used to conduct the parameter learning for $\theta$ and $\phi$.
	
	\subsection{Actor Network}
	For the policy netowrk of the actor, we employ the typical attention based seq2seq framework to conduct the summary generation. Long Short-Term Memory (LSTM) \cite{hochreiter1997long} or Gated Recurrent Unit (GRU) \cite{cho2014learning} can be used as the cell for RNN.  Considering that GRU has comparable performance but with less parameters and more efficient computation, we employ GRU as the basic recurrent model which updates the variables according to the following operations:
	\begin{equation}
		\begin{array}{l}
			\mathbf{r}_t = \sigma (\mathbf{W}_{xr}\mathbf{x}_t + \mathbf{W}_{hr}\mathbf{h}_{t - 1} + \mathbf{b}_r)\\
			\mathbf{z}_t = \sigma (\mathbf{W}_{xz}\mathbf{x}_t + \mathbf{W}_{hz}\mathbf{h}_{t - 1} + \mathbf{b}_z)\\
			\mathbf{g}_t = \tanh (\mathbf{W}_{xh}\mathbf{x}_t + \mathbf{W}_{hh}(\mathbf{r}_t \odot \mathbf{h}_{t - 1}) + \mathbf{b}_h)\\
			\mathbf{h}_t = \mathbf{z}_t \odot \mathbf{h}_{t - 1} + (1 - \mathbf{z}_t) \odot \mathbf{g}_t
		\end{array}
	\end{equation}
	where $\mathbf{r}_t$ is the reset gate, $\mathbf{z}_t$ is the update gate.
	$\odot$ denotes the element-wise multiplication. $tanh$ is the  hyperbolic tangent activation function.
	
	We use bidirectional recurrent neural networks as the encoder.
	Let $\mathbf{x}_t$ be the word embedding vector of the $t$-th word in the input source sequence.
	GRU maps $\mathbf{x}_t$ and the previous hidden state $\mathbf{h}_{t-1}$ to the current hidden state $\mathbf{h}_t$ in feed-forward direction and back-forward direction respectively:
	\begin{equation}
		\begin{array}{l}
			{{\mathord{\buildrel{\lower3pt\hbox{$\scriptscriptstyle\rightharpoonup$}} 
						\over {\mathbf{h}}} }_t} = GRU({x_t},{{\mathord{\buildrel{\lower3pt\hbox{$\scriptscriptstyle\rightharpoonup$}} 
						\over {\mathbf{h}}} }_{t - 1}})\\
			{{\mathord{\buildrel{\lower3pt\hbox{$\scriptscriptstyle\leftharpoonup$}} 
						\over {\mathbf{h}}} }_t} = GRU({x_t},{{\mathord{\buildrel{\lower3pt\hbox{$\scriptscriptstyle\leftharpoonup$}} 
						\over {\mathbf{h}}} }_{t - 1}})
		\end{array}
	\end{equation}
	Then the final hidden state $\mathbf{h}_t^e \in \mathbb{R}^{2k_h}$ is concatenated using the hidden states from the two directions:
	\begin{equation}
		{\mathbf{h}}_t^e = {{\mathord{\buildrel{\lower3pt\hbox{$\scriptscriptstyle\rightharpoonup$}} 
					\over {\mathbf{h}}} }_t}||\mathord{\buildrel{\lower3pt\hbox{$\scriptscriptstyle\leftharpoonup$}} 
			\over {\mathbf{h}}}_t 
	\end{equation}
	
	The decoder is also a GRU based recurrent neural network with improved attention modeling. The first hidden state $\mathbf{h}_1^d$ of the decoder is initialized using the average of all the source input hidden states: 
	\begin{equation}
		\mathbf{h}_1^d = \frac{1}{{{m}}}\sum\limits_{t = 1}^{{m}} {\mathbf{h}_t^e}
	\end{equation}
	where $\mathbf{h}_t^e$ is the source input hidden state. $m$ is the input sequence length.
	Then the two layers of GRUs are designed to conduct the attention weights calculation and decoder hidden states update.
	On the first layer, the hidden state is calculated only using the current input word embedding $\mathbf{y}_{t-1}$ and the  previous hidden state $\mathbf{h}_{t-1}^{d_1}$:
	\begin{equation}
		\mathbf{h}_t^{d_1} = GRU_1(\mathbf{y}_{t-1}, \mathbf{h}_{t-1}^{d_1})
	\end{equation}
	where the superscript $d_1$ denotes the first decoder GRU layer.
	Then the attention weights at the time step $t$ are calculated based on the relationship of $\mathbf{h}_t^{d_1}$ and all the source hidden states $\{\mathbf{h}_t^e\}$. Let $a_{i,j}$ be the attention weight between $\mathbf{h}_i^{d_1}$ and $\mathbf{h}_j^{e}$, which can be calculated using the following formulation:
	\begin{equation}
		\setlength{\abovedisplayskip}{3pt}
		\setlength{\belowdisplayskip}{3pt}
		\begin{aligned}
			{a_{i,j}} &= \frac{{\exp ({e_{i,j}})}}{{\sum\nolimits_{j' = 1}^{{T^e}} {\exp ({e_{i,j'}})} }}\\
			{e_{i,j}} &= {\mathbf{v}^T}\tanh (\mathbf{W}_{hh}^d\mathbf{h}_i^{{d_1}} + \mathbf{W}_{hh}^e\mathbf{h}_j^e + {\mathbf{b}_a})
		\end{aligned}
	\end{equation}
	where $\mathbf{W}_{hh}^d \in \mathbb{R}^{k_h \times k_h}$, $\mathbf{W}_{hh}^e \in \mathbb{R}^{k_h \times 2k_h}$, $\mathbf{b}_a \in \mathbb{R}^{k_h}$, and $\mathbf{v} \in \mathbb{R}^{k_h}$.
	The attention context is obtained by the weighted linear combination of all the source hidden states:
	\begin{equation}
		\setlength{\abovedisplayskip}{3pt}
		\setlength{\belowdisplayskip}{3pt}
		{\mathbf{c}_t} = \sum\nolimits_{j' = 1}^{{T^e}} {{a_{t,j'}}\mathbf{h}_{j'}^e} 
	\end{equation}
	
	The final hidden state $\mathbf{h}_t^{d_2}$ is the output of the second decoder GRU layer, jointly considering the word $\mathbf{y}_{t-1}$, the previous hidden state $\mathbf{h}_{t-1}^{d_2}$, and the attention context $\mathbf{c}_t$:
	\begin{equation}
		\mathbf{h}_t^{d_2} = GRU_2(\mathbf{y}_{t-1}, \mathbf{h}_{t-1}^{d_2}, \mathbf{c}_t)
	\end{equation}
	Then the probability of generating any target word $y_t$ is given as follows:
	\begin{equation}
		\hat{\mathbf{y}}_t = \varsigma({\mathbf{W}_{hy}^{d}}\mathbf{h}_{t}^{d_2} + {\mathbf{b}^{d}_{hy}})
	\end{equation}
	where ${\mathbf{W}_{hy}^{d}} \in \mathbb{R}^{k_y \times k_h}$ and ${\mathbf{b}^{d}_{hy}} \in \mathbb{R}^{k_y}$. $\varsigma(\cdot)$ is the softmax function.
	In the prediction state, we use the beam search algorithm \cite{koehn2004pharaoh} for decoding and generating the best summary.
	
	\subsection{Critic \RN{1}}
	We employ the traditional maximum likelihood estimator as the policy assessment criteria for Critic \RN{1}. For convenience and consistency, we represent the value $V{_\theta}$ using negative log likelihood (NLL). Given the ground truth summary
	${Y} = \{\mathbf{y}_1, \mathbf{y}_2, \ldots, \mathbf{y}_n\}$ for the input sequence $X$, we can depict $V{_\theta}$ as:
	\begin{equation}
		{V_\theta(\hat Y, X) } = \sum\limits_{t = 1}^n { - \log p({y_t}|{y_{<t}},X)}
	\end{equation}
	Since negative log likelihood is differentiable, we can update the actor parameter $\theta$ using gradient method directly: 
	\begin{equation}
		\label{eq:c1}
		\theta  = \theta  - \alpha_{\RN{1}} \nabla {V_\theta(\hat Y, X)}
	\end{equation}
	where $\alpha_{\RN{1}}$ is the learning rate using Critic \RN{1}.
	\subsection{Critic \RN{2}}
	
	Although teacher forcing training strategy \cite{williams1989learning} used in the actor can accelerate the training convergence, it leads to the discrepancy problem between training and testing. Moreover, no summary sentence level assessment criteria are considered to guarantee the summary quality. 
	Thus we design another critic named global summary quality estimator to relieve this problem.
	The global summary quality estimator is a neural network based binary classifier which can be regarded as a discriminator between generated summaries and the ground truth. The aim is to make the generated summaries indistinguishable from the human-written ones. Hence it will output low scores for the low-quality summaries and high scores for the ground truth.
	We denote the estimator as $V_{\phi}(\bar Y, X)$ where $X$ is the input source sequence, and $\bar Y$ represent the ground truth $Y$ or the Monte-Carlo sequence $\hat{Y}$ sampled from the policy $p(Y|X)$ for one action.
	
	For the representation of $X$, we employ the original states from the actor encoder component by concatenating the last hidden states of the outputs of the bidirectional recurrent neural networks for $X$:
	\begin{equation}
		{\mathbf{h}}^x = {{\mathord{\buildrel{\lower3pt\hbox{$\scriptscriptstyle\rightharpoonup$}} 
					\over {\mathbf{h}^x_n}} }}||\mathord{\buildrel{\lower3pt\hbox{$\scriptscriptstyle\leftharpoonup$}} 
			\over {\mathbf{h}^x_1}} 
	\end{equation}
	where ${{\mathord{\buildrel{\lower3pt\hbox{$\scriptscriptstyle\rightharpoonup$}} 
				\over {\mathbf{h}^x_n}} }}$ and $\mathord{\buildrel{\lower3pt\hbox{$\scriptscriptstyle\leftharpoonup$}} 
		\over {\mathbf{h}^x_1}} $ are the two final states to represent the sentence in two directions respectively.
	For $\bar Y$, we propose another bidirectional recurrent neural networks for the representation learning and obtain the states similarly with $X$:
	$
	{\mathbf{h}}^y = {{\mathord{\buildrel{\lower3pt\hbox{$\scriptscriptstyle\rightharpoonup$}} 
				\over {\mathbf{h}^y_n}} }}||\mathord{\buildrel{\lower3pt\hbox{$\scriptscriptstyle\leftharpoonup$}} 
		\over {\mathbf{h}^y_1}} 
	$.
	Then we add a non-linear transformation to combine $\mathbf{h}^x$ and $\mathbf{h}^y$:
	\begin{equation}
		{\mathbf{h}^c} = \tanh (\mathbf{W}_{xh}^c{\mathbf{h}^x} + \mathbf{W}_{yh}^c{\mathbf{h}^y} + {\mathbf{b}^c})
	\end{equation}
	where $\mathbf{W}_{xh}^c \in \mathbb{R}^{k_h \times k_h}$, $\mathbf{W}_{yh}^c \in \mathbb{R}^{k_h \times k_h}$, and $\mathbf{b}^c \in \mathbb{R}^{k_h}$. Finally, we add a softmax layer to let the model output a binary category variable:
	\begin{equation}
		{\mathbf{v}^c} = \varsigma (\mathbf{W}_{hv}^c{\mathbf{h}^c} + {\mathbf{b}^v})
	\end{equation}
	where $\mathbf{W}_{hv}^c \in \mathbb{R}^{k_h \times 2}$ and $\mathbf{b}^v \in \mathbb{R}^{2}$.
	We have mentioned that we treat the ground truth as the positive instance and the sampled sequence as the negative instance. So we directly let the first dimension of $\mathbf{v}^c$ represent the positive label.
	Therefor we can depict $V_{\phi}(\bar Y, X)$ as follows:
	\begin{equation}
		V_{\phi}(\bar Y, X) = \mathbf{v}^c_{[0]}
	\end{equation}
	which represents the assessment value given the source text $X$ and a sequence $\bar Y$ needs to be critiqued.
	
	We use cross entropy as the loss function $J(\phi)$ which used to judge the classification performance and optimize the parameter $\phi$ of Critic \RN{2}:
	\begin{equation}
		\phi  = \phi  - \alpha_{\phi} \nabla {J(\phi) }
	\end{equation} 
	
	We use policy gradient method to employ $V_{\phi}(\bar Y, X)$ to conduct the optimization of the policy parameter $\theta$ for the actor.
	
	\textbf{Policy gradient}. Each word $\hat y_t$ in the generated summary $\hat Y$ is sampled from the action space via the corresponding probability distribution $\hat{\mathbf{y}}$, which is a non-differentiable operation.
	Thus the assessment value obtained from $V_{\phi}(\hat Y, X)$ can not be propagated to update the policy parameters of the actor effectively.
	After investigations, we find that both REINFORCE \cite{williams1992simple} and Gumbel-Softmax \cite{jang2017categorical} can be used to tackle the problem. We employ REINFORCE in our work, considering that it can be integrated into our actor-critic training framework naturally, as well as the better performance mentioned in some other tasks \cite{li2017adversarial,wu2017adversarial,yu2017seqgan}.
	
	More specifically, we regard $V_{\phi}(\hat Y, X)$ as the reward function for one episode of summary generation, and the objective is to maximize the expected reward: 
	\begin{equation}
		J(\theta ) = {E_{\hat Y \sim p(Y|X)}}{V_\varphi }(\hat Y,X)
	\end{equation}  
	Then the the policy gradient can be estimated using the REINFORCE trick, i.e., the likelihood ratio \cite{williams1992simple}:
	\begin{equation}
		\nabla J(\theta ) = \sum\limits_{t = 1}^T {\nabla \log p({y_t}|{y_{ < t}},X,\theta ) \cdot {V_\varphi }(\hat Y,X)} 
	\end{equation}  
	The policy parameter $\theta$ is updated using the obtained gradient:
	\begin{equation}
		\theta  = \theta  - \alpha_{\RN{2}} \nabla {J(\theta) }
	\end{equation} 
	
	\subsection{Alternating Actor-Critic Training}
	We propose an alternating training strategy to train the actor and the critic. The details are shown in Algorithm~\ref{alg:ac}. As mentioned in several previous works \cite{li2017adversarial,wu2017adversarial,yu2017seqgan,bahdanau2017actor,fedus2018maskgan}, we also find that it is difficult to make the training stable if the joint training is conducted from the start point. Therefore, at the beginning, we also conduct a pre-training state to optimize the policy network of the actor using Critic \RN{1}, i.e., the maximum likelihood estimator, to update the parameters according to Equation~\ref{eq:c1}.
	After that, we conduct the joint training for the actor the critic.
	However, we find that only relying on $V_{\phi}(\hat Y, X)$ of Critic \RN{2} to train the policy network will make the model generate worse results. Moreover, although a weighted combination of the criteria of Critic \RN{1} and \RN{2}, as did in \cite{paulus2017deep}, can improve the performance, we still need to tune the weight parameter carefully. In our framework, we propose an alternating training strategy.
	More specifically, the parameter $\phi$ for Critic \RN{2} is updated every $K_3$ iterations. Then, as shown in Algorithm~\ref{alg:ac}, we update the policy parameter $\theta$ twice using different assessment values from Critic \RN{1} and \RN{2} respectively. 
	Actually, from a different perspective, we can regard the training procedure using Critic \RN{2} as a \textbf{fine-tuning} component for the traditional typical attention based seq2seq framework. Thus, the alternating training strategy can accelerate the convergence rate, make the training more stable, and achieve better prediction performance.

	\begin{algorithm}[!t]
		\caption{Actor-critic training framework.}
		\label{alg:ac}
		\small
		\begin{algorithmic}[1]
			\REQUIRE The actor network with parameter $\theta$ and the critic network with parameter $\phi$. The training dataset $(\mathcal{X}, \mathcal{Y})$.
			\ENSURE The best summarization model $\theta^*$.
			\STATE Initialize $\theta$ and $\phi$;
			\WHILE{Not Converged}
			\STATE \# pre-traing the actor using Critic \RN{1}:
			\FOR{$i \in \{1, \dots, K_1\}$}
			\STATE $\theta  = \theta  - \alpha_{\RN{1}} \nabla {V_\theta(\hat Y, X)}$
			\ENDFOR
			\STATE \# actor-critic training:
			\FOR{$i \in \{1, \dots, K_2\}$}
			\STATE \# optimize Critic \RN{2} every $K_3$ steps
			\IF { $i \ \ \% \ \ K_3 = 0$}
			\STATE sample $(X, Y)$ from real data as the positive instance;
			\STATE sample $\hat Y \sim G_{\theta}(\cdot | X)$ as the negative instance;
			\STATE optimize Critic \RN{2}: \\
			\ \ \ \ $\phi  = \phi  - \alpha_{\phi} \nabla {J(\phi) }$
			\ENDIF
			\STATE \# alternating training:
			\STATE train the actor using Critic \RN{1}: \\
			\ \ \ \ $\theta  = \theta  - \alpha_{\RN{1}} \nabla {V_\theta(\hat Y, X)}$
			\STATE train the actor using Critic \RN{2}: \\
			\ \ \ \ $\theta  = \theta  - \alpha_{\RN{2}} \nabla {J(\theta) }$
			\ENDFOR
			\ENDWHILE
			\RETURN $\theta^* = \theta$.
		\end{algorithmic}
	\end{algorithm}
	
	\section{Experimental Setup}
	
	\subsection{Datesets}
	We train and evaluate our framework on three popular benchmark datasets.
	\textbf{Gigawords} is an English sentence summarization dataset prepared based on Annotated Gigawords\footnote{https://catalog.ldc.upenn.edu/ldc2012t21} by extracting the first sentence from news reports with the headline to form a source (the first sentence)-summary (headline) pair.
	We directly download the prepared dataset  used in \cite{rush2015neural}.
	It roughly contains 3.8M training pairs, 190K validation pairs, and 2,000 test pairs.The test set is identical to the one used in all the comparative baseline methods.
	\textbf{DUC-2004}\footnote{http://duc.nist.gov/duc2004} is another English dataset only used for testing in our experiments. It contains 500 documents. Each document contains 4 model summaries written by experts. The length of the summary is limited to 75 bytes.
	\textbf{LCSTS} is a large-scale Chinese short text summarization dataset, consisting of pairs of (short text, summary) collected from Sina Weibo\footnote{http://www.weibo.com} \cite{hu2015lcsts}.
	We take Part-I as the training set, Part-II as the development set, and Part-III as the test set. There is a score in the range of $1\sim5$ labeled by human to indicate how relevant an article and its summary is. We only make use of those pairs with scores no less than 3. The size of the three sets are 2.4M, 8.7k, and 725 respectively.
	In our experiments, we only take Chinese character sequence as input, without performing word segmentation.

	\subsection{Evaluation Metrics}
	We use ROUGE score \cite{lin2004rouge} as our evaluation metric with standard options.
	The basic idea of ROUGE is to count the number of overlapping units between generated summaries and the reference summaries, such as overlapped n-grams, word sequences, and word pairs.
	F-measures of ROUGE-1 (R-1), ROUGE-2 (R-2) and ROUGE-L (R-L) are reported.
	
	\subsection{Comparative Methods}
	We compare our model \textbf{AC-ABS} with some baselines and state-of-the-art methods.
	Since the datasets are quite standard, so we just extract the results from their papers. Therefore the baseline methods on different datasets may be slightly different.
	\textbf{TOPIARY} \cite{zajic2004bbn} is the best on DUC2004 Task-1 for compressive text summarization.	It combines a system using linguistic based transformations and an unsupervised topic detection algorithm for compressive text summarization.
	\textbf{MOSES+} \cite{rush2015neural} uses a phrase-based statistical machine translation system trained on Gigaword to produce summaries.
	It also augments the phrase table with ``deletion'' rules to improve the baseline performance, and MERT is also used to improve the quality of generated summaries.
	\textbf{ABS} and \textbf{ABS+} \cite{rush2015neural} are both the neural network based models with local attention modeling for abstractive sentence summarization.
	ABS+ is trained on the Gigaword corpus, but combined with an additional log-linear extractive summarization model with handcrafted features.
	\textbf{RNN} and \textbf{RNN-context} \cite{hu2015lcsts} are two seq2seq architectures. RNN-context integrates attention mechanism to model the context.
	\textbf{CopyNet} \cite{gu2016incorporating} integrates a copying mechanism into the sequence-to-sequence framework.
	\textbf{RNN-distract} \cite{chen2016distraction} uses a new attention mechanism by distracting the historical attention in the decoding steps.
	\textbf{RAS-LSTM} and \textbf{RAS-Elman} \cite{chopra2016abstractive} both consider words and word positions as input and use convolutional encoders to handle the source information.
	For the attention based sequence decoding process, RAS-Elman selects Elman RNN \cite{elman1990finding} as decoder, and RAS-LSTM selects Long Short-Term Memory architecture \cite{hochreiter1997long}.
	\textbf{LenEmb} \cite{kikuchi2016controlling} uses a mechanism to control the summary length by considering the length embedding vector as the input.
	\textbf{ASC+FSC$_1$} \cite{miao2016language} uses a generative model with attention mechanism to conduct the sentence compression problem.
	The model first draws a latent summary sentence from a background language model, and then subsequently draws the observed sentence conditioned on this latent summary.
	\textbf{lvt2k-1sent} and \textbf{lvt5k-1sent} \cite{nallapati2016abstractive} utilize a trick to control the vocabulary size to improve the training efficiency.
	\textbf{SEASS} \cite{zhou2017selective} integrates a selective gated network into the seq2seq framework to control the information flow from encoder to decoder.
	\textbf{DRGD} \cite{li2017deep} proposes a deep recurrent generative decoder to enhance the modeling ability of latent structures in the target summaries.
	\textbf{GBN} \cite{chen2018generative} proposes a generative bridging network in which a bridge module is introduced to assist the training of the sequence prediction model.

	\subsection{Experimental Settings}
	For the parameters used to control the training iterations, we let $K_1 = 5$ (epochs), $K_2 = 2$ (epochs), and $K_3 = 50$ (iterations).
	For the experiments on the English dataset Gigawords, we set the dimension of word embeddings to 300, and the dimension of hidden states and latent variables to 500.
	The maximum length of documents and summaries is 100 and 50 respectively.
	The batch size of mini-batch training is 256.
	For DUC-2004, the maximum length of summaries is 75 bytes.
	For the dataset of LCSTS, the dimension of word embeddings is 350.
	We also set the dimension of hidden states and latent variables to 500.
	The maximum length of documents and summaries is 120 and 25 respectively, and the batch size is also 256. 
	The beam size of the decoder was set to be 10.
	Adadelta \cite{zeiler2012adadelta} with hyperparameter $\rho = 0.95$ and $\epsilon = 1e-6$ is used for gradient based optimization.
	Although adadelta has a adaptive learning rate, we sill set the learning rate $\alpha_{\phi} = \alpha_{\RN{1}} = \alpha_{\RN{2}} = 0.1$ during the last two epochs of alternating training.  
	Our neural network based framework is implemented using Theano \cite{2016arXiv160502688short}.


	\section{Results and Discussions}
	
	\subsection{Training Analysis}
	
	\begin{figure*}[!t]
		\begin{subfigure}{0.25\textwidth}
			\includegraphics[width=\linewidth]{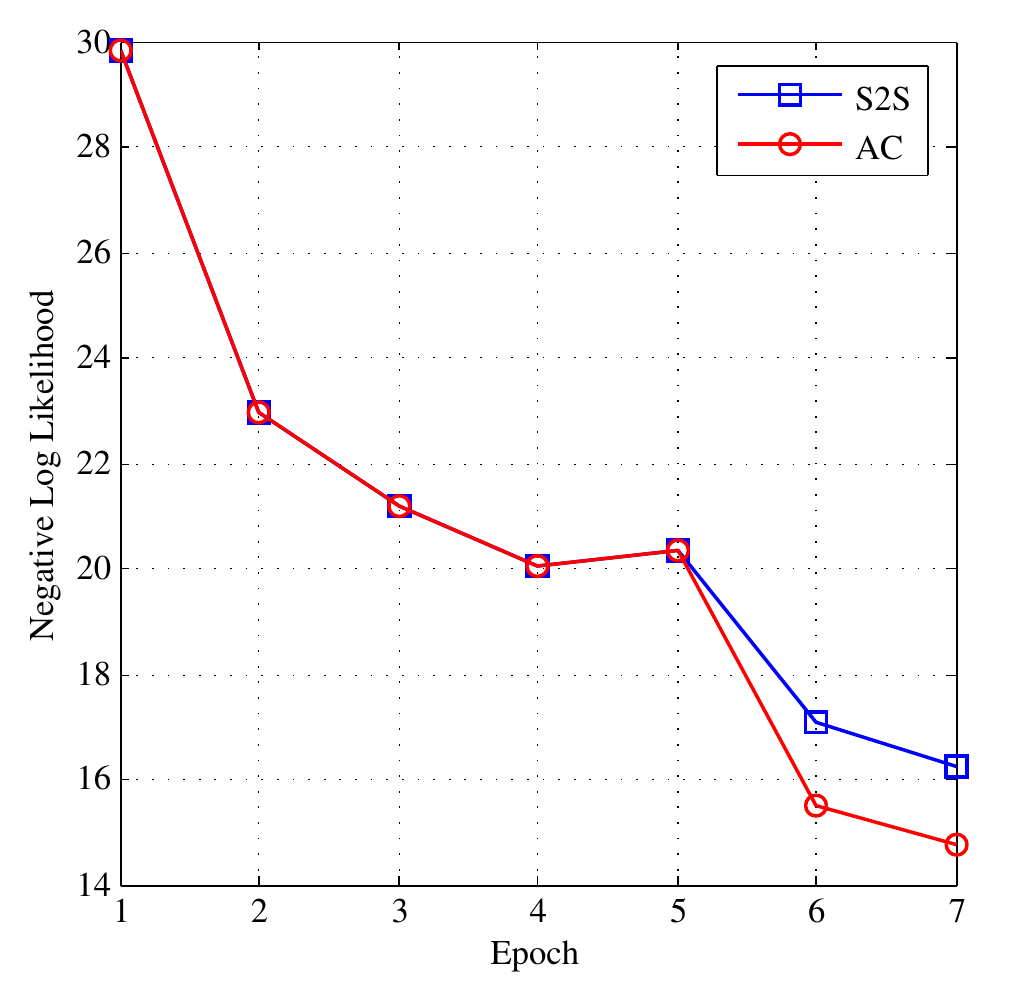}
			\caption{Negative log likelihood} \label{fig:nll}
		\end{subfigure}\hspace*{\fill}
		\begin{subfigure}{0.25\textwidth}
			\includegraphics[width=\linewidth]{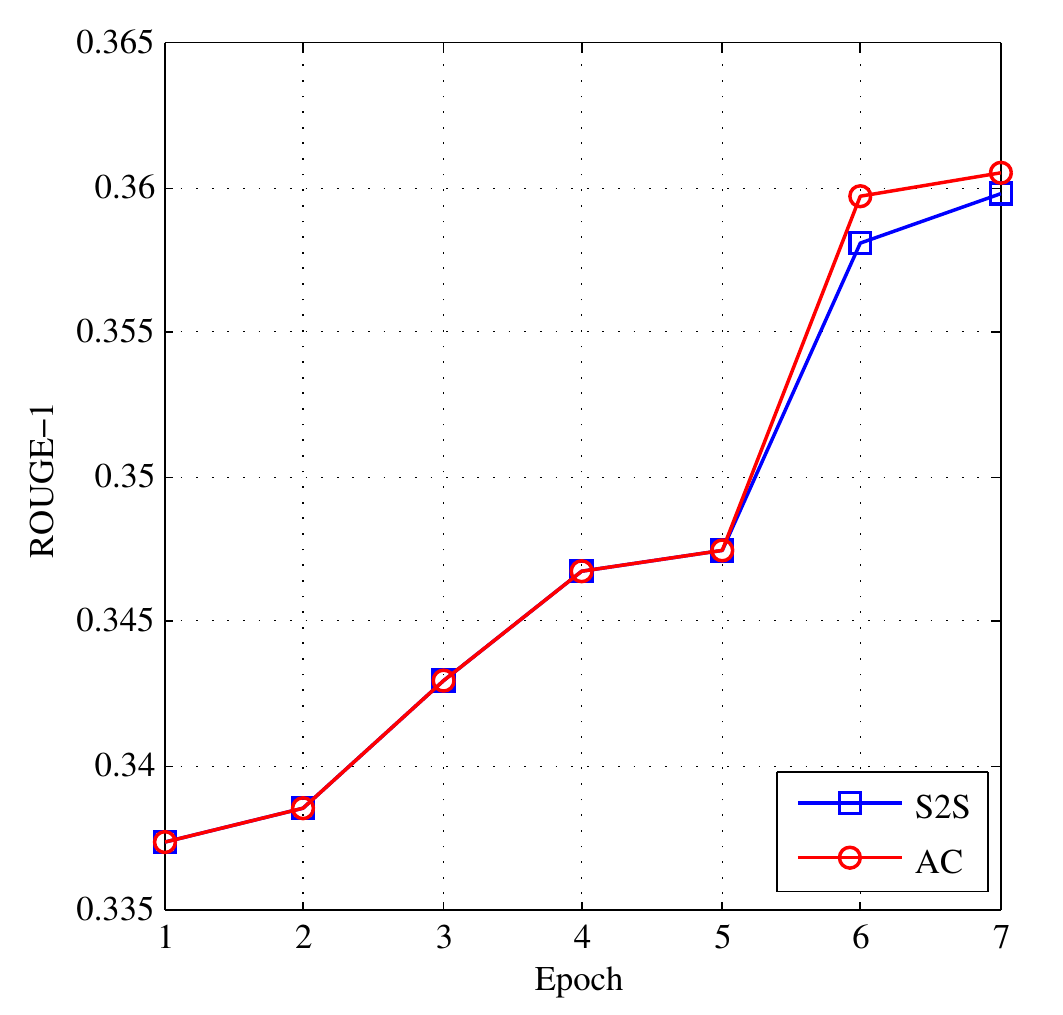}
			\caption{ROUGE-1} \label{fig:r1}
		\end{subfigure}\hspace*{\fill}
		\begin{subfigure}{0.25\textwidth}
			\includegraphics[width=\linewidth]{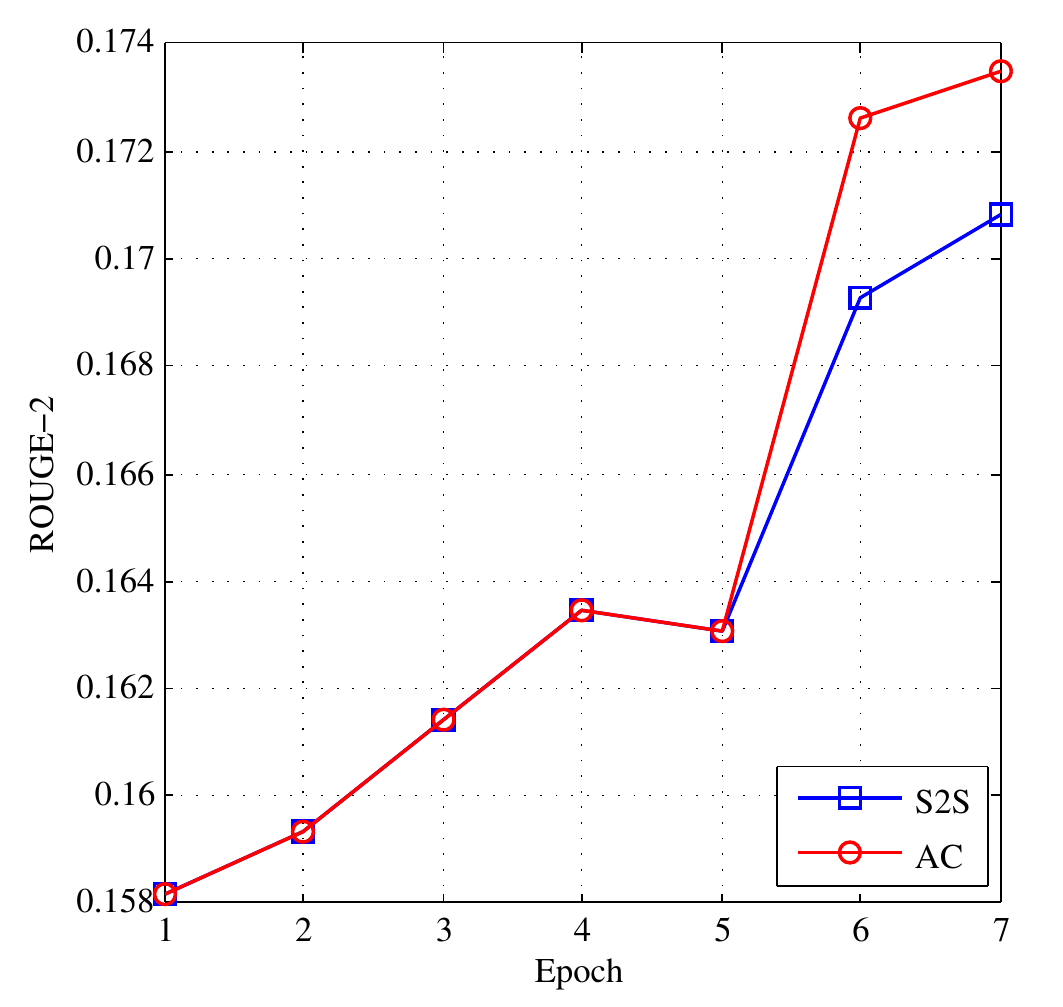}
			\caption{ROUGE-2} \label{fig:r2}
		\end{subfigure}\hspace*{\fill}
		\begin{subfigure}{0.25\textwidth}
			\includegraphics[width=\linewidth]{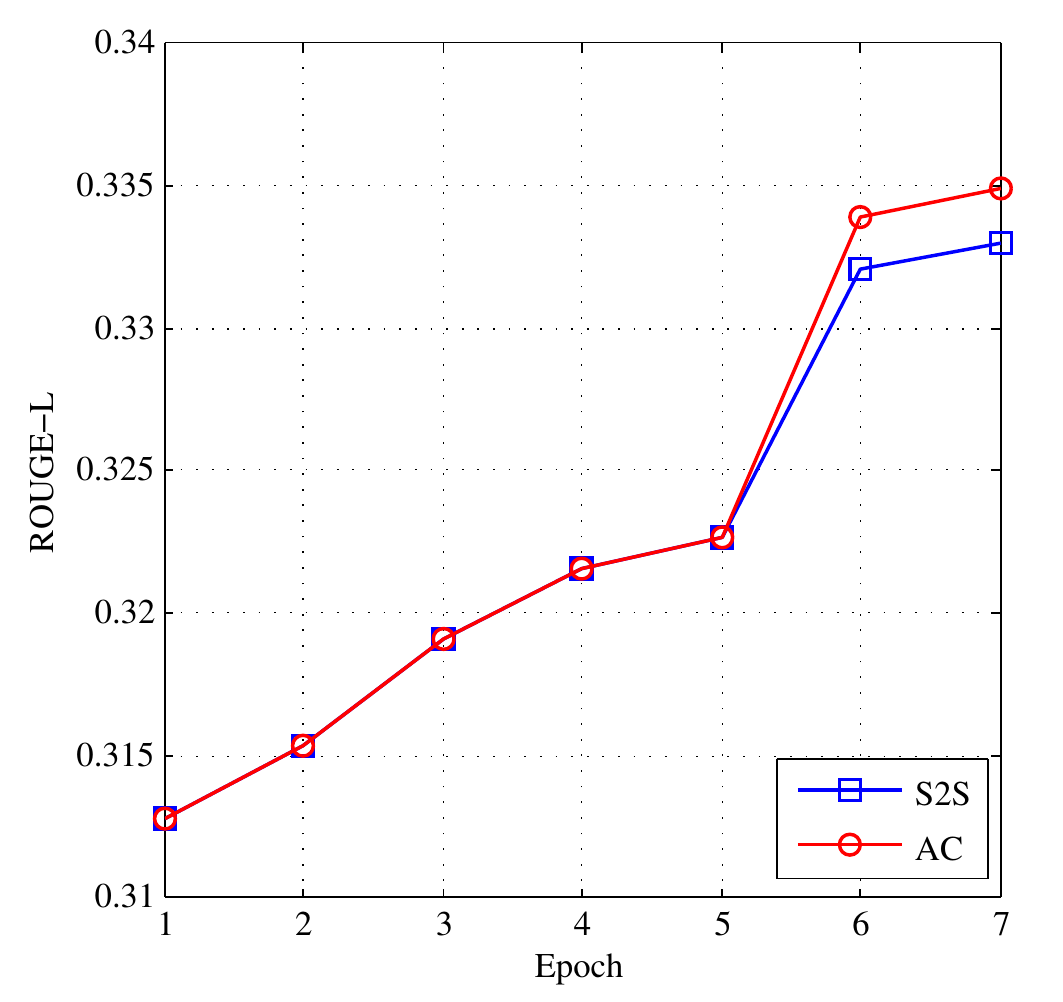}
			\caption{ROUGE-L} \label{fig:rl}
		\end{subfigure}
		\caption{The performance on the metric of likelihood and ROUGE evaluation for the two different training strategies: the alternating actor-critic (AC) training strategy and the attention based seq2seq (S2S) model. It illustrates that the alternating actor-critic training strategy can indeed accelerate the training convergence as well as improve the prediction performance.} \label{fig:training}
	\end{figure*}
	
	To illustrate the effectiveness of the actor-critic training strategy, we conduct a comparison on the metrics of negative log likelihood (NLL) and ROUGE between our proposed training methods and the traditional seq2seq framework.
	The results are shown in Figure~\ref{fig:training}. Epoch 1$\sim$5 is the pre-training state. From epoch 6, we start the actor-critic training strategy.
	From Figure~\ref{fig:training}(a), we can see that our actor-critic training strategy can accelerate the convergence and obtain a better solution with lower loss value. To verify whether the model is over-fitting or not, we provide the ROUGE evaluation on the validation dataset. Figure~\ref{fig:training}(b-d) show that all ROUGE-1, ROUGE-2, and ROUGE-L have been improved, which illustrates that our training framework can indeed improve the prediction performance. 
	
	\begin{table}[!t]
		\centering
		\caption{Objective values during the actor-critic training.}
		\label{tab:lossd}
		\begin{tabular}{c c c c}
			\hline
			\textbf{Loss}  & \textbf{Epoch 0} & \textbf{Epoch 1} & \textbf{Epoch 2} \\
			\hline
			$J_{\phi}$       & 0.4950 & 0.0425 & 0.0066  \\
			$-J_{\theta}$     & 0.5531 & 0.0823 & 0.0137  \\
			\hline
		\end{tabular}
	\end{table}
	
	Moreover, according to Table~\ref{tab:lossd}, we find that the training procedure for Critic \RN{2} can achieve a very small objective value, which means that Critic \RN{2} can discriminate the generated summary with the ground truth easily. So it is very difficult to let the critic cheat the actor. This is the reason why we mentioned that we prefer to name our framework with ``actor-critic training'' rather than ``adversarial training''.
	
	\subsection{ROUGE Evaluation}
	
	\begin{table}[!t]
		\centering
		\caption{ROUGE-F1 on Gigawords}
		\label{tab:rouge-agiga}
		\begin{tabular}{p{2.6cm} c c c}
			\hline
			\textbf{System}  & \textbf{R-1} & \textbf{R-2} & \textbf{R-L}  \\
			\hline
			ABS       & 29.55 & 11.32 & 26.42  \\
			ABS+       & 29.78 & 11.89 & 26.97  \\
			RAS-LSTM       & 32.55 & 14.70 & 30.03  \\
			RAS-Elman       & 33.78 & 15.97 & 31.15  \\
			ASC-FSC$_1$       & 34.17 & 15.94 & 31.92  \\
			lvt2k-1sent     & 32.67 & 15.59 & 30.64  \\
			lvt5k-1sent     & 35.30 & 16.64 & 32.62  \\
			GBN    & 35.26 & 17.22 & 32.67  \\
			\textbf{AC-ABS}       & \textbf{36.05} & \textbf{17.35} & \textbf{33.49}  \\
			\hline
		\end{tabular}
	\end{table}

	\begin{table}[!t]
		\centering
		\caption{ROUGE-Recall on DUC2004}
		\label{tab:rouge-duc04}
		\begin{tabular}{p{2.6cm} c c c}
			\hline
			\textbf{System}  & \textbf{R-1} & \textbf{R-2} & \textbf{R-L} \\
			\hline
			TOPIARY & 25.12 & 6.46 & 20.12  \\
			MOSES+ & 26.50 & 8.13 & 22.85  \\
			ABS       & 26.55 &	7.06 &	22.05  \\
			ABS+       & 28.18 & 8.49 & 23.81  \\
			RAS-Elman       & 28.97 & 8.26 & 24.06  \\
			RAS-LSTM       & 27.41 & 7.69 & 23.06  \\
			LenEmb       & 26.73 & 8.39 & 23.88  \\
			lvt2k-1sen       & 28.35 & 9.46 & 24.59  \\
			lvt5k-1sen & 28.61 & 9.42 & 25.24  \\
			SEASS       & 29.21 & 9.56 & 25.51 \\
			DRGD       & 28.99 & 9.72 & 25.28 \\
			\textbf{AC-ABS}       & \textbf{29.41} & \textbf{9.84} & \textbf{25.85} \\
			\hline
		\end{tabular}
		\vspace{-2mm}
	\end{table}

	\begin{table}[!t]
		\centering
		\caption{ROUGE-F1 on LCSTS}
		\label{tab:rouge-lcsts}
		\begin{tabular}{p{2.6cm} c c c}
			\hline
			\textbf{System}  & \textbf{R-1} & \textbf{R-2} & \textbf{R-L} \\
			\hline
			RNN       & 21.50 & 8.90 & 18.60  \\
			RNN-context       & 29.90 & 17.40 & 27.20  \\
			CopyNet       & 34.40 & 21.60 & 31.30  \\
			RNN-distract & 35.20 & 22.60 & 32.50  \\
			DRGD       & 36.99 & 24.15 & 34.21 \\
			\textbf{AC-ABS}       & \textbf{37.51} & \textbf{24.68} & \textbf{35.02} \\
			\hline
		\end{tabular}
		\vspace{-2mm}
	\end{table}
	
	The results on the English datasets of Gigawords and DUC-2004 are shown in Table~\ref{tab:rouge-agiga} and Table~\ref{tab:rouge-duc04} respectively.
	Our model actor-critic based abstractive summarization framework (AC-ABS)  achieves the best performance on all the ROUGE metrics. 
	
	The basic summarization model in our framework is still an attention-based seq2seq framework, so we design the comparative experiments mainly with the model in typical attention based seq2seq model.
	lvt2k-1sent and lvt5k-1sent \cite{nallapati2016abstractive} provide a strong baseline in their paper. 
	As shown in Table~\ref{tab:rouge-agiga}, our framework performs better than lvt5k-1sent, which means that the actor-critic training strategy can indeed improve the prediction performance of the typical seq2seq summarization models.   
	It is worth noting that the methods lvt2k-1sent and lvt5k-1sent utilize linguistic features such as parts-of-speech tags, named-entity tags, and TF and IDF statistics of the words as part of the document representation.
	Generally, more useful features can indeed improve the performance. Nevertheless our framework is still better than them.
	
	The results on the Chinese dataset LCSTS are shown in Table~\ref{tab:rouge-lcsts}.
	Our model AC-ABS also achieves the best performance. Although CopyNet employs a copying mechanism to improve the summary quality, RNN-distract considers attention information diversity in their decoders, and DRGD integrates a recurrent variational auto-encoder into the typical seq2seq framework, our model is still better than these methods demonstrating that the effectiveness of the alternating actor-training strategy.
	We also believe that integrating the copying mechanism and coverage diversity in our framework will further improve the summarization performance. 
	
	
	\subsection{Summary Case Analysis}
	
	\begin{table}[!t]
		\small
		\centering
		\caption{Examples of the generated summaries.}
		\label{tab:cases}
		\begin{tabular}{p{7.2cm}}
			\hline
			\hline
			\textbf{S(1)}: japan 's toyota team europe were banned from the world rally championship for one year here on friday in a crushing ruling by the world council of the international automobile federation fia.\\
			\textbf{Golden}: toyota are banned for a year.\\
			\textbf{seq2seq}: toyota 's world rally europe banned from world rally championship.\\
			\textbf{AC-ABS}: \textbf{toyota team europe banned for one year.}\\
			\hline
			\textbf{S(2)}: president  's springer spaniel , spot , was old and ailing.\\
			\textbf{Golden}: former head usher at white house witnessed first families at their most vulnerable.\\
			\textbf{seq2seq}: president 's spaniel spaniel spaniel spaniel becomes secretary of s springer spaniel spaniel spaniel spaniel spaniel spaniel.\\
			\textbf{AC-ABS}: \textbf{us president s springer spaniel becomes new president of the united states.}\\
			\hline
			\textbf{S(3)}: u.s. home resales posted the largest monthly increase in at least \#\# years last month as first-time buyers rushed to take advantage of a tax credit that expires this fall.\\
			\textbf{Golden}: july home sales surge more than \# percent.\\
			\textbf{seq2seq}: u.s. home resales rose \#.\# \% to \#.\#\#\# bln rate in \#\# years.\\
			\textbf{AC-ABS}: \textbf{u.s. home sales post largest monthly rise in \#\# years.}\\
			\hline
			\textbf{S(4)}: the thai government has set aside \#\#\# million baht about \#\#.\#\# million u.s. dollars to support new eco-tourism plans during \#\#\#\#-\#\#\#\# , according to a report of the thai news agency tna tuesday.\\
			\textbf{Golden}: thai government to support eco-tourism.\\
			\textbf{seq2seq}: thailand to support new eco-tourism in \#\#\#\#-\#\#\#\#.\\
			\textbf{AC-ABS}: \textbf{thailand to support new eco-tourism plans.}\\
			\hline
			\hline
		\end{tabular}
	\end{table}
	
	To illustrate the effectiveness of our proposed actor-critic training framework for abstractive summarization vividly, we compare the generated summaries by AC-ABS and the typical attention based seq2seq framework used in some other works such as \cite{chopra2016abstractive,li2017deep}.
	The source texts, golden summaries, and the generated summaries are shown in Table~\ref{tab:cases}. 
	In S(1), the seq2seq result is ``toyota 's world rally europe banned from world rally championship.'', which contains phrase ``world rally'' twice. The seq2seq summary in S(2)  contains multiple term ``spaniel''.
	It is obvious that our framework AC-ABS generate better summaries for S(1) and S(2) respectively. The summaries generated by seq2seq for S(3) and S(4) contain more unimportant noisy symbols such as ``\#''. Because the critic component will assign a small value for this kind of sentences during training, so our framework is able to avoid of generating too much noisy symbols. For example, the seq2seq result is ``thailand to support new eco-tourism in \#\#\#\#-\#\#\#\#.'', and we can see that AC-ABS generate a better sentence ``thailand to support new eco-tourism plans.'' by removing the symbol sequence.

	\section{Conclusions}
	We present a training framework for neural abstractive summarization based on actor-critic approaches. For the actor, we employ a typical attention based seq2seq framework as the policy network. For the critic, we combine the maximum likelihood estimator with a global summary quality estimator. An alternating training strategy is proposed to conduct the joint learning of the actor and critic models. Extensive experiments on some benchmark datasets in different languages show that our framework achieves improvements over the state-of-the-art methods. 
	

\bibliography{emnlp2018}
\bibliographystyle{acl_natbib_nourl}
\end{document}